\pdfoutput=1

\documentclass[11pt]{article}
\usepackage[final]{acl}
\usepackage{ulem}
\usepackage{algorithm}
\usepackage{algpseudocode}
\usepackage{amssymb}

\usepackage{times}
\usepackage{latexsym}

\usepackage[T1]{fontenc}

\usepackage[utf8]{inputenc}

\usepackage{microtype}

\usepackage{multirow} 
\usepackage{booktabs} 
\usepackage{amssymb}
 \usepackage{amsmath}
\usepackage{multicol}
\usepackage{inconsolata}
\usepackage{array}

\usepackage{graphicx}
\usepackage{tabularx}
\usepackage{pdfpages}

\title{GDLLM: A Global Distance-aware Modeling Approach Based on Large Language Models for Event Temporal Relation Extraction}

\author{
  \textbf{Jie Zhao\textsuperscript{1}},
  \textbf{Wanting Ning\textsuperscript{2}},
  \textbf{Yuxiao Fei\textsuperscript{1}},
  \textbf{Yubo Feng\textsuperscript{1}},
  \textbf{Lishuang Li\textsuperscript{1}\thanks{Corresponding author}}  
  \\  
  \textsuperscript{1}School of Computer Science and Technology, Dalian University of Technology, China
  \\  
  \textsuperscript{2}Luddy School of Informatics, Computing, and Engineering, Indiana University Indianapolis, the United States
  \\  
  \texttt{zj18835245681@mail.dlut.edu.cn},
  \texttt{ningw@iu.edu},
  \\
  \texttt{fyx2379@mail.dlut.edu.cn},
  \texttt{argmax@126.com},
  \texttt{lils@dlut.edu.cn}
  \\
}

\begin{document}
\maketitle

\begin{abstract} In Natural Language Processing(NLP), Event Temporal Relation Extraction (ETRE) is to recognize the temporal relations of two events. Prior studies have noted the importance of language models for ETRE. However, the restricted pre-trained knowledge of Small Language Models(SLMs) limits their capability to handle minority class relations in imbalanced classification datasets. For Large Language Models(LLMs), researchers adopt manually designed prompts or instructions, which may introduce extra noise, leading to interference with the model's judgment of the long-distance dependencies between events. To address these issues, we propose \textbf{GDLLM}, a \textbf{G}lobal \textbf{D}istance-aware modeling approach based on \textbf{LLM}s. We first present a distance-aware graph structure utilizing Graph Attention Network(GAT) to assist the LLMs in capturing long-distance dependency features. Additionally, we design a temporal feature learning paradigm based on soft inference to augment the identification of relations with a short-distance proximity band, which supplements the probabilistic information generated by LLMs into the multi-head attention mechanism. Since the global feature can be captured effectively, our framework substantially enhances the performance of minority relation classes and improves the overall learning ability. Experiments on two publicly available datasets, TB-Dense and MATRES, demonstrate that our approach achieves state-of-the-art (SOTA) performance. 

\end{abstract}

\section{Introduction}
\begin{figure}[t]
  \centering\includegraphics[width=0.9\columnwidth]{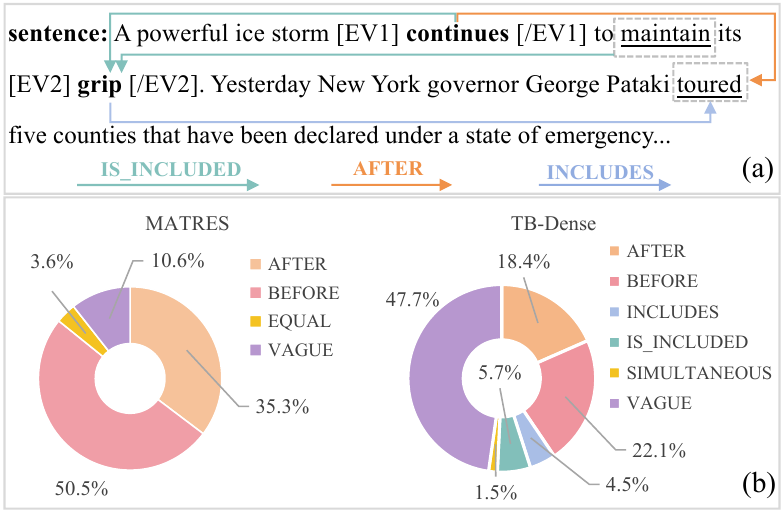}
  \caption{(a) is an example of the ETRE task. Above the arrows in the legend are the corresponding relation categories. ``\([EV_i]\)'' is the hand-crafted symbol that can explicitly mark event boundaries in such examples. (b) is the relation distribution on two datasets.}
  \label{fig:sample}
\end{figure}

In Natural Language Processing (NLP), Event Temporal Relation Extraction (ETRE) aims to identify temporal connections between event pairs. As illustrated in Figure~\ref{fig:sample}(a), in the given sentence, the relation between the target
Event1 ‌\textbf{continues} and Event2 ‌\textbf{grip} is \textit{IS\_INCLUDED}. 


Much of the existing studies pay attention to the crucial role of language models for ETRE, especially Small Language Models(SLMs). Some research utilizes SLMs to form certain rules for temporal realtion\citep{zhang:2022,man：2022,zhuang:2023}. Prior SOTA model MulCo\citep{yao:2024} combines GNNs and the model of BERT variants via multi-scale knowledge distillation to enhance the performance of ETRE. However, the restricted pre-trained knowledge of SLMs limits their capability to handle minority class relations in imbalanced classification datasets\citep{uzzaman:2013,guan:2021}. Although some researchers have invested substantial effort in it \citep{han:2019,ning:2024,yuan:2024}, the performance of their models is still suboptimal on two popular datasets, MATRES\citep{ning:2019} and TB-Dense\citep{cassidy:2014}. As depicted in Figure~\ref{fig:sample}(b), the relation ``\textit{SIMULTANEOUS}'' that refers to two events happening simultaneously only takes 1.5\% in the TB-Dense dataset, while ``\textit{VAGUE}'' has 47.7\%\citep{yuan:2024}.  

Recent advancements have noted the impressive capabilities of Large Language Models(LLMs) for ETRE. However, based on the powerful learning ability for contextual knowledge, prior studies rely on manually designed prompts and instructions to fine-tune LLMs\citep{hu:2025,xu:2025}, leading to noise accumulation\citep{chen:2024,Shi:2024,diao:2025} that interferes with the model’s judgment of the global event relation feature.  
As shown in Figure~\ref{fig:sample}(a), unlike most event pairs among Events ``\textbf{continues}'', ``\textbf{grip}'' and ``\textbf{toured}'', two another events occur between Events ``\textbf{continues}'' and ``\textbf{toured}'' in the text and make their distance of the occurrence order is longer. This indicates that there are two different event relation features that constitute the global feature: long-distance dependency and short-distance proximity band. Since modeling global event relation feature poses a challenge for researchers\citep{shi:2025}, they often neglect the recognition of long-distance dependency features when adopting manually designed prompts and instructions, which is also not conducive to handling minority class relations in imbalanced classification datasets.

To resolve the aforementioned problems, we propose \textbf{GDLLM}, a \textbf{G}lobal \textbf{D}istance-aware modeling approach based on \textbf{LLM}s, enabling the effective identification of event relations with the global feature to alleviate the impact of data imbalance on classification results. To be specific, we select the Graph Attention Network(GAT) to assist the fine-tuned LLMs in capturing event relations with long-distance dependency features, which circumvents the limitations of manually designed prompts or instruction templates.  Compared to the ``hard classification'' (0/1 decision labels) as graph edge features, we integrate the probability distribution generated by LLMs into GAT  to learn more comprehensive relation information. Both the probabilistic information and the multi-head attention mechanism augment the identification of relations with a short-distance proximity band. Since the global feature can be captured effectively, our framework substantially enhances the performance of minority relation classes and the overall learning ability.

Our contributions can be summarized as follows:
\begin{itemize}
    \item We propose \textbf{GDLLM}, a \textbf{G}lobal \textbf{D}istance-aware modeling approach based on \textbf{LLM}s. Specifically, we introduce a global modeling method integrating LLMs and GAT, which is to identify the minority categories more effectively in imbalanced classification datasets.
    \item We present a distance-aware graph structure utilizing Graph Attention Network to assist the fine-tuned LLMs in capturing event relations with long-distance dependency features, which circumvents the limitations imposed by manually designed prompts or instructions.
    \item We design a temporal feature learning paradigm based on soft inference to augment the relation extraction with a short-distance proximity band. Rather than 0/1 decision labels, the probability distribution we selected as edge features enables the GAT to learn more comprehensive relation information. 
    \item We conduct extensive experiments on two public datasets, TB-Dense and MATRES, which demonstrate that our approach outperforms all existing LLM-based and GNN-based benchmarks, achieving state-of-the-art (SOTA) performance without manually designed prompts or instructions for LLMs.

\end{itemize}


\begin{figure*}[t]
\centering \includegraphics[width=0.95\textwidth,height=0.8\textheight,keepaspectratio]{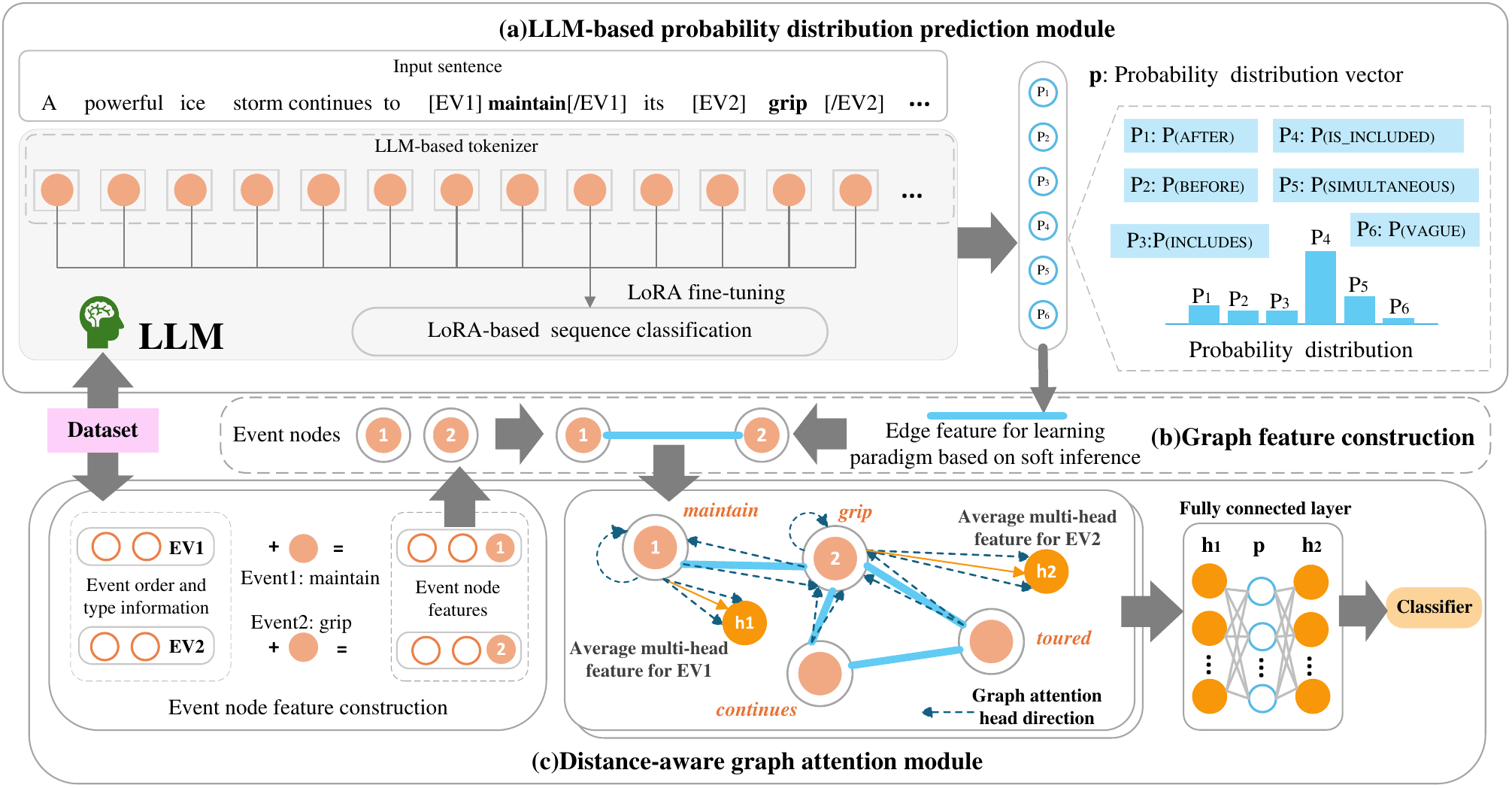}
  \caption{Overall architecture of our proposed method.}
  \label{fig: model}
\end{figure*}

\section{Method}

In this section, we introduce the overall architecture of our proposed \textbf{GDLLM} method, which is depicted in Figure~\ref{fig: model}. Firstly, we formulate the ETRE task. Secondly, we introduce the LLM-based probability distribution prediction module. Thirdly, we present the distance-aware graph attention module for long-distance feature capture, and a temporal feature learning paradigm for short-distance feature learning. Finally,  our framework will enhance the performance of minority relation classes in the imbalanced classification dataset by capturing the global feature.

\subsection{Problem Formulation}
Following previous work, we define ETRE as a text classification task. Given a sentence \(T\) that contains two events \(E_1\) and \(E_2\), our aim is to identify the temporal relation between these two events. The output of our model is the particular temporal relation label prediction.

\subsection{Probability Distribution Prediction}
\textbf{Input and fine-tuning for the LLM.}
In our work, the unified format defined that input to LLMs from datasets contains manually designed symbols of the form \([EV_i]\) in the given sentence \(T\), where $i$ denotes the ordinal number of an event pair. It serves as a marker to annotate the boundaries of the event. 
Before generating probabilistic information, we fine-tune the LLM based on LoRA\citep{hu:2022}. As depicted in Figure~\ref{fig: model}(a), the LoRA fine-tuned technique is used for sequence classification, which is adopted for parameter-efficient fine-tuning. 



\textbf{Probability generation.}
 As shown in Figure~\ref{fig: model}(a), while applying the LoRA fine-tuning, the model is ready to make predictions of probabilistic information generated by the LLM to construct edge features of graph structure, forming the soft inference-based temporal feature learning paradigm we designed. For each pair of events \((E_i, E_j)\) in the document, the LLM outputs a probability distribution over a set of predefined and annotated event relation classes. 
 
 Specifically, we define \(c\) as the number of event relation classes, and the output of the LLM for the event pair \((E_i, E_j)\) is a vector \(\mathbf{p}_{ij} \in \mathbb{R}^c\). In the inference process of LoRA tuning, the model first generates a set of logits for each event pair. These logits are then passed through the softmax function. This operation converts the logits into probabilities, which represent the likelihood of each event pair belonging to different relation labels. For \(c\) kinds of relation types, and a specific event pair \((E_i, E_j)\), the probability of it belonging to relation \(r\) is denoted as \(P(p = r|E_i, E_j)\). Mathematically, the logits for an event pair are \(z_1, z_2, \cdots, z_n\), then the probability \(P(p = r|E_i, E_j)\) is calculated as:
\begin{equation}
P(p = r|E_i, E_j)=\frac{e^{z_r}}{\sum_{n = 1}^{c}e^{z_n}},
\end{equation}
which normalizes the logits so that the sum of probabilities for all relation classes is equal to 1, and they are all stored in the probability distribution vectors \(\mathbf{p}_{ij} \in \mathbb{R}^C\). As depicted in Figure~\ref{fig: model}(a), rather than determining the most likely temporal relation between events, these probabilities are made to be a vector sequence distribution to provide more comprehensive pre-trained information for the subsequent module. For the TB-Dense dataset, which is shown in Figure~\ref{fig: model}(a) as an instance, the LLM provides the prediction distribution of the six labels it has, while the MATRES dataset does so for the four labels it possesses.


\begin{figure*}[t]

\centering\includegraphics[width=\linewidth]{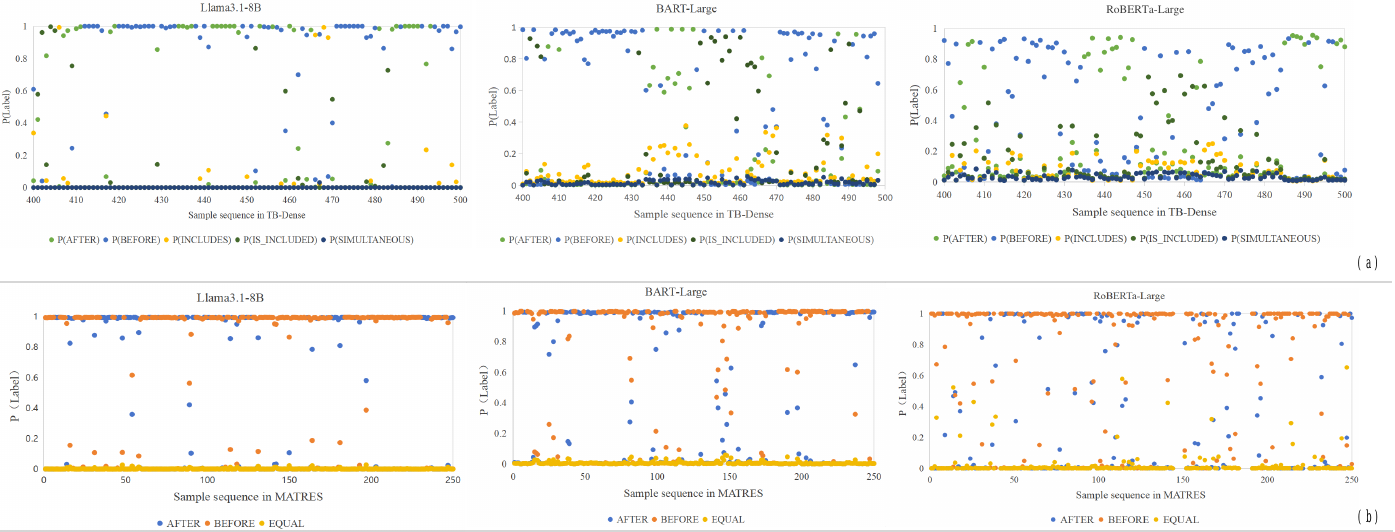}
  \caption{The distribution of probabilities generated by different language models on the TB-Dense (a) and the MATRES (b) datasets.}
  \label{fig:P}
\end{figure*}

The training objective is the cross-entropy loss for multi-class classification based on LLM, which does not participate in the final loss calculation, and the calculation details of the loss function are similar to the final classification.

Notably, a topic that deserves discussion is why we choose LLMs to be the main language models. We generate probability distributions through different language models and visualize these distributions in scatter plots for comparison of the accuracy of the probability value. To be specific, we randomly select a sample sequence from the TB-Dense and MATRES datasets, respectively. Then, we compare the distribution of probabilities for positive samples. It can be seen from Figure~\ref{fig:P} that LLMs can present probability values with higher accuracy, which always assign a value closer to 1 for the relation category with the highest probability, while for low probability prediction values, their distributions tend to be closer to zero.


\subsection{Distance-aware Graph Attention Module}
From the previous section, we have obtained the probability distribution vectors \(\mathbf{p}_{ij} \in \mathbb{R}^C\) for event pair predictions generated by the LLM. Next, we will introduce the construction of graph features first, followed by the temporal feature learning paradigm based on soft inference. Since our graph structure is a solution for capturing relation features at different distances, we define the proposed GAT-based architecture as a distance-aware approach.

\textbf{Graph feature construction.}
As depicted in Figure~\ref{fig: model}(c), we construct a graph to model the relations between events based on every complete document. Compared with traditional graph construction methods, our approach aims to be more conducive to enabling the graph structure to learn accurate global relational features at an earlier stage. Each event \(E_i\) in the document and its order and type information are both represented as a node \(v_i\in V\). And the node features \(h_i^{(0)}\in\mathbb{R}^{d_h}\) are obtained from the dataset corresponding to the event. 

For edge feature, which is shown as Figure~\ref{fig: model}(b), it exists between every pair of nodes, and the edge features are initialized as the probability distribution vectors \(\mathbf{p}_{ij} \in \mathbb{R}^C\) for the event pair \((E_i,E_j)\), which is to form our temporal feature learning paradigm based on soft inference.

\begin{table*}

\centering
\begin{tabular}{lccccccc}
\toprule
\multirow{2}{*}{\centering \textbf{Model}} & \multirow{2}{*}{\textbf{Language Model}} & \multicolumn{3}{c}{\textbf{TB-Dense}} & \multicolumn{3}{c}{\textbf{MATRES}} \\
\cmidrule(lr){3-5} \cmidrule(lr){6-8}
 &  & P(\%) & R(\%) & F1(\%) & P(\%) & R(\%) & F1(\%) \\
\midrule
TIMERS*\citep{mathur:2021} & BERT-Base  & 48.1 & 65.2 & 67.8 & 81.1 & 84.6 & 82.3 \\
SGT*\citep{zhang:2022} & BERT-Large & - & - & 67.1 & - & - & 80.3 \\
RSGT*\citep{zhou:2022} & RoBERTa-Base & 68.7 & 68.7 & 68.7 &  82.2 & 85.8 & 84.0 \\
Bayesian \citep{tan:2023}  & BART-Large & - & - & 65.0 & 79.6 & 86.0 & 82.7 \\
Unified \citep{huang:2023} & RoBERTa-Large & - & - & 68.1 & - & - & 82.6 \\
TCT \citep{ning:2024} & BART-Large & 70.3 & 71.6 & 70.9 & 79.0 & 87.2 & 82.9 \\
CPTRE \citep{yuan:2024} & BERT-Base & 73.4 & 69.5 & 71.4 & 81.3 & 86.3 & 84.2 \\
MulCo* \citep{yao:2024} & RoBERTa-Large & - & - & 85.6 & - & - & 90.4 \\
MAQInstruct \citep{xu:2025} & Llama2-7B & - & - & - & 85.5 & 83.9 & 84.7 \\
LLMERE \citep{hu:2025} & Llama3.1-8B & - & - & - & 82.6 & 88.7 & 85.5 \\
 \midrule
 SLM(BART) with GAT&  BART-Large & 75.8 & 68.9  & 71.3 & 80.6 & 84.7 &  81.2\\
SLM(RoBERTa) with GAT & RoBERTa-Large & 70.8 & 68.8  & 69.2 & 82.4 & 91.7 & 86.4 \\
\midrule
 GDLLM\_Qwen(Ours) &  Qwen2.5-7B & 85.3 & 86.5 & 86.1 & \textbf{86.8\textbf} & 94.8 & 90.6\\
 \textbf{GDLLM(Ours)} &  \textbf{Llama3.1-8B} & \textbf{88.3} & \textbf{86.6} & \textbf{87.5} & 86.5 & \textbf{95.9} & \textbf{90.9} \\

\bottomrule
\end{tabular}
\caption{The overall experiment results on the two datasets. Models marked with a * use the GNN-based approach. The F1 score means micro-F1.}
\label{tab:results}
\end{table*}

\textbf{Temporal feature learning paradigm.}
We design a temporal feature learning paradigm based on soft inference as depicted in Figure~\ref{fig: model}(b), which is to supplement the probabilistic information generated by LLMs into the multi-head attention mechanism. This paradigm shifts the edge feature representation from the previous 0/1 decision label to a probability distribution for “soft inference”, which augments the identification of relations with a short-distance proximity band. To achieve this, we apply this paradigm to the edge feature learning of GAT, which constructs a graph structure to model event relations with a multi-head attention mechanism.

Specifically, our Graph Attention Network architecture consists of multiple layers of multi-head attention mechanisms. In our implementation, in order to enable the model to learn more diverse feature combinations and interaction information, we adopt two layers with \(K=8\) attention heads. In the first GAT Layer, for each node \(v_i\), the output of the \(k\)-th attention head is computed as:
\begin{equation}
\hat{\mathbf{h}}_{i,k}^{(1)}=\sigma\left(\sum_{j\in\mathcal{N}(i)}\alpha_{ij,k}\mathbf{W}_k^{(1)}\mathbf{h}_j^{(0)}\right),
\end{equation}
where \(\mathcal{N}(i)\) is the set of neighboring nodes of \(v_i\), \(\mathbf{W}_k^{(1)}\in\mathbb{R}^{d_h\times d_{h1}}\) is the weight matrix for the \(k\) - th head, \(\sigma\) is the activation function LeakyReLU, and the attention coefficients \(\alpha_{ij,k}\) are calculated as:
\begin{equation}
    \mathbf{z}_{ij,k} = \mathbf{a}_k^{\top}[\mathbf{W}_k^{(1)}\mathbf{h}_i^{(0)}\parallel\mathbf{W}_k^{(1)}\mathbf{h}_j^{(0)}\parallel\mathbf{p}_{i,j}],
\end{equation}
\begin{equation}
     \mathbf{z}_{im,k} = \mathbf{a}_k^{\top}[\mathbf{W}_k^{(1)}\mathbf{h}_i^{(0)}\parallel\mathbf{W}_k^{(1)}\mathbf{h}_m^{(0)}\parallel\mathbf{p}_{i,m}],  
\end{equation}
\begin{equation}
    \alpha_{ij,k}=\frac{\exp\left(\text{LeakyReLU}(\mathbf{z}_{ij,k})\right)}{\sum_{m}\exp\left(\text{LeakyReLU}(\mathbf{z}_{im,k})\right)},
\end{equation}
where \(\mathbf{a}_k\in\mathbb{R}^{3d_{h1}}\) is a learnable attention vector, ``\(\parallel\)'' denotes concatenation, and $m\in\mathcal{N}(i)$. Afterwards, the output of the first layer for node \(v_i\) is then concatenated with the outputs of all heads: $\mathbf{h}_i^{(1)}=\text{Concat}(\hat{\mathbf{h}}_{i,1}^{(1)},\cdots,\hat{\mathbf{h}}_{i,K}^{(1)})$. For the second GAT layer, following a similar process of the first layer, we get the output of the \(k\)-th attention head $\hat{\mathbf{h}}_{i,k}^{(2)}$, and the final average multi-head feature $\mathbf{h}_i^{(2)}$.

\textbf{Final Classification.} In the final classification stage, as depicted in Figure~\ref{fig: model}(c), we integrate the output of the second GAT layer and the processed edge features $\mathbf{p}_{i,j}$. We concatenate these two types of features as: $\mathbf{h}_{o}=[\mathbf{h}_i^{(2)}\parallel\mathbf{p}_j\parallel\mathbf{h}_j^{(2)}]$. The concatenated feature vector $\mathbf{h}_{o}$ is then fed into a fully-connected layer. The output of the fully-connected layer is calculated as follows:
\begin{equation}
\mathbf{s}=\mathbf{W}_{\text{cls}}\mathbf{h}_{o}+\mathbf{b}_{\text{cls}},
\end{equation}
where $\mathbf{W}_{\text{cls}}\in\mathbb{R}^{d_{h2}\times C}$ is the weight matrix of the classification layer, and $\mathbf{b}_{\text{cls}}\in\mathbb{R}^{C}$ is the bias vector of the classification layer.

Subsequently, we apply the softmax function for the output of the fully-connected layer to obtain the predicted probability distribution over the classes. The softmax function is defined as:
\begin{equation}
\hat{\mathbf{y}}=\text{softmax}(\mathbf{s}),
\end{equation}
where $\hat{\mathbf{y}}$ is the predicted probability for an event pair $(E_i, E_j)$.

We employ the cross-entropy loss function to measure the difference between the final predicted probability and the true label. Given the true label $\mathbf{y}=(y_1, y_2, \cdots, y_C)$, the cross-entropy loss for this event pair is calculated as:
\begin{equation}
\mathcal{L}(\mathbf{y}, \hat{\mathbf{y}})=-\sum_{k = 1}^{C}y_k\log(\hat{y}_k).
\end{equation}



\section{Experiments and Results}

\subsection{Datasets and Metrics}
We validate our approach on two widely adopted datasets: MATRES \citep{ning:2019}, and TB-Dense\citep{cassidy:2014}. In accordance with prior study\citep{ning:2024,huang:2023}, we adopt the micro-F1 score, with the \textit{VAGUE} label excluded, as the evaluative metric for the datasets. Our data splits statistics also follow the previous studies\citep{ning:2024,huang:2023}.

\subsection{Experimental Setup}
We compare our method with the following baselines: \textbf{1) LLM-based approaches:} these methods leverage LLMs to encode contextual information and perform temporal reasoning through prompt or instruction tuning\citep{xu:2025,hu:2025}. Prior studies also explore zero-shot temporal relation extraction using different prompt strategies\citep{yuan:2023,xu:2025}. Following this work, we conduct zero-shot experiments on two LLMs, the closed-source GPT4o\citep{hurst:2024} and the open-source Llama3.1. \textbf{2) Graph-based approaches:} These models construct event graphs to capture temporal information, often using Graph Neural Networks (GNNs) to propagate information\citep{mathur:2021,zhang:2022,zhou:2022,yao:2024}. \textbf{3) Other benchmarks:} Methods that do not fit into the above categories but have shown strong performance, often combining neural networks or heuristic features\citep{huang:2023,tan:2023,ning:2024,yuan:2024}. In addition, we employ RoBERTa-Large \citep{liu:2019} and BART-Large \citep{lewis:2020} as two baseline models for comparison of SLMs.

As for fine-tuning LLMs, the LoRA rank is set to 16. All experiments are trained on NVIDIA A800 GPUs with 80GB of memory. In this paper, following previous work for hyperparameter optimization\citep{yao:2024}, we employ the HEBO (Heuristic-Efficient Bayesian Optimization) algorithm.


 
  
       



 


\subsection{Main Results}
As shown in Table~\ref{tab:results}, our method achieves SOTA performance in all existing methods and baseline methods. It is apparent from this table that very few models utilize LLMs as their language model to chase superior performance, but our methods adopt LLMs and outperform all previous models without manually designed prompts or instructions tuning. Meanwhile, unlike previous approaches, we arrange LLMs not as a standalone reasoning model, which also shows the effectiveness of utilizing distance-aware graph structure to form our approach, and is validated to capture the global feature of temporal event relations.

Additionally, our method \textit{GDLLM(Ours)} outperforms the previous SOTA model\citep{yao:2024} that only adopts SLM as its language model, achieving an increase of 1.9\% on the micro-F1 comparison of the TB-Dense dataset. Although the relatively smaller data scale of the MATRES dataset and its characteristic of extremely imbalanced class distribution may limit the model's ability to fully learn the event categories, our method still effectively outperforms the existing best result by 0.5\%. This not only validates that our temporal feature learning paradigm based on distance-aware modeling enables the model to learn global features with different proximity more effectively, but also indicates the impressive capabilities of the LLM we employed. We also observe significant advantages of our method compared with the two SLMs we developed, the RoBERTa-Large and BART-Large. That is because, compared with large language models, SLMs have certain limitations in the volume of pre-training data and only generate less accurate probability distribution prediction vectors.

\begin{figure}[t]
 \includegraphics[width=0.95\linewidth]{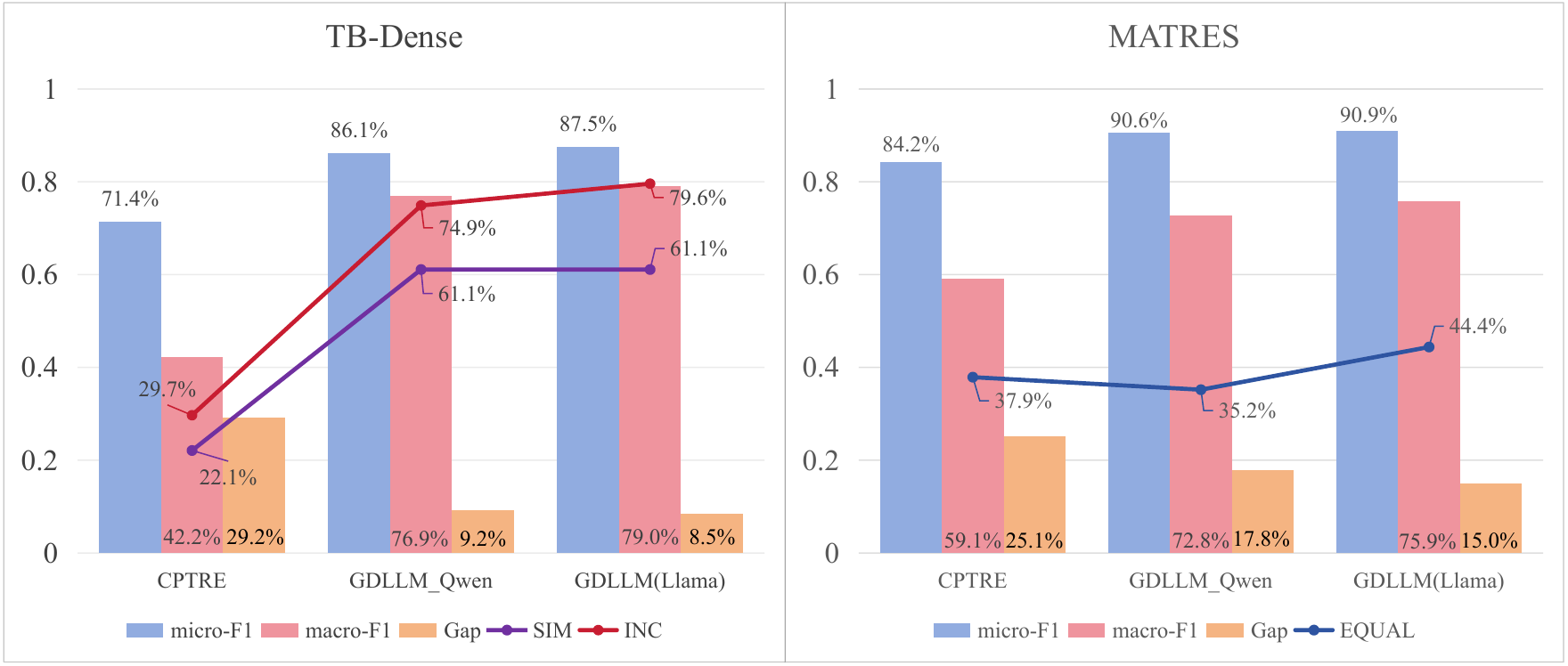}
  \caption{The performance of micro-F1, macro-F1, and the F1 score of some minority categories between our methods and the selected study. SIM: \textit{SIMULTANEOUS}. INC: \textit{INCLUDES}. Gap: the difference between micro-F1 and macro-F1. A lower Gap value indicates better performance of the model on minority categories.}
  \label{fig:gap}
\end{figure}

\subsection{Performance on Minority Categories}
To confirm that our method is valid to identify the minority categories more effectively in the situation of imbalanced data, we also compare the micro-F1 score and macro-F1 score between our methods and the most recent study on a similar issue\citep{yuan:2024}, which reports that their results outperform earlier studies on the macro-F1 score. According to respective definitions, macro-F1 gives equal weight to each category, while micro-F1 gives equal weight to each sample. This ensures that if a model achieves a severe gap between micro and macro, the model cannot perform well on minority categories. 

It can be seen from the data in Figure~\ref{fig:gap} that the ``Gap'' scores on our methods are obviously lower than those in the model CPTRE. In general, our GDLLM (Llama3.1) model outperforms CPTRE on all minority categories. Although our method GDLLM\_Qwen performs suboptimally on the \textit{EQUAL} class when using Qwen as the language model, we think that is because the \textit{EQUAL} class has an exceptionally low count, causing the model's severely biased prediction on the categories with extremely high proportions during training. On the basis of the analysis above, our proposed model achieves significantly better performance on all datasets regarding macro-F1 scores, and it indeed improves the model’s performance on minority temporal relation classes.


\begin{table}

  \centering
 
  \begin{tabular}{cccccc}
    \toprule
     \textbf{Method} & \textbf{LLMs} & \textbf{P(\%)} & \textbf{R(\%)} & \textbf{F1(\%)} \\
    \midrule
  
   \textbf{GDLLM} &  Llama3.1  & \textbf{86.5} & \textbf{95.9} & \textbf{90.9}  \\
        w/o LP & - & 64.6 & 73.4 & 68.7 \\
        w/o GD & Llama3.1 & 77.2 & 79.0 & 78.1 \\
         w/o PI & Llama3.1 & 78.9 & 86.7 & 82.6 \\
      \midrule
       
     \textbf{GDLLM}  & Qwen2.5 & \textbf{86.8} & \textbf{94.8} & \textbf{90.6} \\
       w/o LP & - & 64.6 & 73.4 & 68.7 \\
       w/o GD  & Qwen2.5 & 74.0  & 82.7  & 77.1 \\
        w/o PI  & Qwen2.5 & 75.3 & 82.1 & 79.5 \\
    \bottomrule
  \end{tabular}

  \caption{The ablation experimental results on the MATRES dataset.  ``\textit{w/o LP}'' only uses the GAT-based multi-head attention mechanism.}
  \label{tab:MAablation_results}
\end{table}

\subsection{Ablation Study}

Table~\ref{tab:MAablation_results} illustrates the ablation experimental results on the MATRES dataset (Table~\ref{tab:TBablation_results} shows ablation results on the TB-Dense dataset. Since the ablation results on the two datasets are equivalent, in this subsection, we analyze the results on the MATRES dataset as a representative case). Our experiments are based on two LLMs (Llama3.1 and Qwen2.5). When analyzing the impact of removing components from the \textbf{GDLLM} method, we observe that ``\textit{w/o \textbf{LP}}'' (without \textbf{L}LM-based \textbf{P}robability Generation), ``\textit{w/o \textbf{GD}}'' (without \textbf{G}AT-based \textbf{D}istance-aware Structure), and ``\textit{w/o \textbf{PI}}'' (without \textbf{P}robabilistic Soft \textbf{I}nference Learning Paradigm) lead to a decrease in performance. 

\begin{table}

  \centering
 
  \begin{tabular}{cccccc}
    \toprule
   \textbf{Method}  & \textbf{LLMs} & \textbf{P(\%)} & \textbf{R(\%)} & \textbf{F1(\%)} \\
    \midrule
  
    \textbf{GDLLM}  & Llama3.1  & \textbf{88.3} & \textbf{86.6} & \textbf{87.5} \\
     w/o LP &  -& 47.3 & 69.1 & 53.2 \\
       w/o GD & Llama3.1 & 67.8 & 58.1 & 62.5 \\
      w/o PI  & Llama3.1 & 62.4 & 72.6 & 66.0 \\
      \midrule
      
     \textbf{GDLLM} & Qwen2.5 & \textbf{85.3} & \textbf{86.5} & \textbf{86.1} \\
     w/o LP & - & 47.3 & 69.1 & 53.2 \\
    w/o GD & Qwen2.5  & 68.0 & 72.7 & 70.8 \\
     w/o PI & Qwen2.5 & 63.6 & 71.5 & 66.0 \\
    \bottomrule
  \end{tabular}

  \caption{The ablation experimental results on the TB-Dense dataset. ``\textit{w/o LP}'' only uses the GAT-based multi-head attention mechanism.}
  \label{tab:TBablation_results}
\end{table}

\textbf{Analysis of \textit{LP}.} As shown in Table~\ref{tab:MAablation_results}.  Through the comparison of the \textit{w/o LP} module, the micro-F1 scores decrease by 22.2\% and 21.9\%, respectively. When the component responsible for generating probability distributions via LLMs is removed, the model consistently achieves the lowest micro-F1 score across all cases. This illustrates that it is challenging for GAT to identify event relations without the probabilistic information generated by LLMs, because the model has been deprived of the powerful capability to capture relation features with a short-distance proximity band. 


\textbf{Analysis of \textit{GD}.} As shown in Table~\ref{tab:MAablation_results}. Comparing the \textit{w/o GD} module, the micro-F1 scores drop by 12.8\% and 13.5\% based on Llama and Qwen, respectively. This indicates the limitation of utilizing LLMs standalone for ETRE, and further demonstrates that our GAT-based distance-aware structure indeed aids the LLMs to better learn the relation features with long-distance dependency.

\textbf{Analysis of \textit{PI}.} As depicted in Table~\ref{tab:MAablation_results}.  We also remove the probabilistic soft inference paradigm for temporal feature learning. That is, we make the LLMs only generate corresponding ``0/1'' label prediction values for edge features, transforming the entire process into a dual-stage hard classification. Comparing the \textit{w/o PI} module, the micro-F1 scores decline by 8.3\% and 11.1\% on the two models. This suggests that enabling the model to learn probabilistic distribution information improves the identification of the event relation of the short-distance proximity band.

Compared with the results in ``\textit{w/o \textbf{PI}}'', the results in ``\textit{w/o \textbf{GD}}'' generally exhibit a more pronounced decline in model performance. This is attributed to the under-learning of a large number of event pairs with long-distance features in the imbalanced dataset, among which the majority belong to minority category labels. 


\begin{table}
\centering
\begin{tabular}{ccccc}
\toprule
  \multirow{2}{*}{\textbf{Method}} & \multicolumn{4}{c}{\textbf{Distance}}  \\
  \cmidrule(lr){2-5} 
  & 2 & 3 & 4 & 5 \\

\midrule

 w/o GD &  \uline{79.3}  & 80.8 & 75.7 &81.8 \\ 

 w/o PI  & 78.1 &  \uline{86.3} & \uline{87.8} & \uline{90.2} \\
   \textbf{Ours} &  \textbf{87.3} & \textbf{93.1} & \textbf{95.7} & \textbf{90.9} \\

\bottomrule
\end{tabular}
\caption{The comparison of micro-F1 scores(\%) of subsets divided based on different distance conditions on the MATRES dataset. The data in bold and with underlines represent the optimal and suboptimal results under each distance condition, respectively.}

\label{tab:long}
\end{table}

\subsection{Performance on Distance Features}
We also test the performance with modules \textit{w/o GD} and \textit{w/o PI} under different distance conditions utilizing Llama3.1. Specifically, we define the distance feature as follows: If there are $n$ other events between the target event pair ($E_i,E_j$), the distance between them is set to 
$n$. As illustrated in Table~\ref{tab:long}, when the distance is progressively increased, the performance of the \textit{w/o GD} models becomes lower than the \textit{w/o PI} models. This indicates that our distance-aware graph structure can more effectively identify temporal relations with longer event distances. Meanwhile, when we remove the \textit{PI} approach, the decline of micro-F1 scores becomes less pronounced as the event distance increases. Notably, when the distance increases to 5, our method only outperforms the model \textit{w/o PI} by 0.7\%. This suggests that the proposed feature learning paradigm based on soft inference can more effectively enhance the performance for events with shorter distances.

\subsection{Analysis of Zero-Shot Experiment}

\begin{figure}[t]


\includegraphics[width=\linewidth]{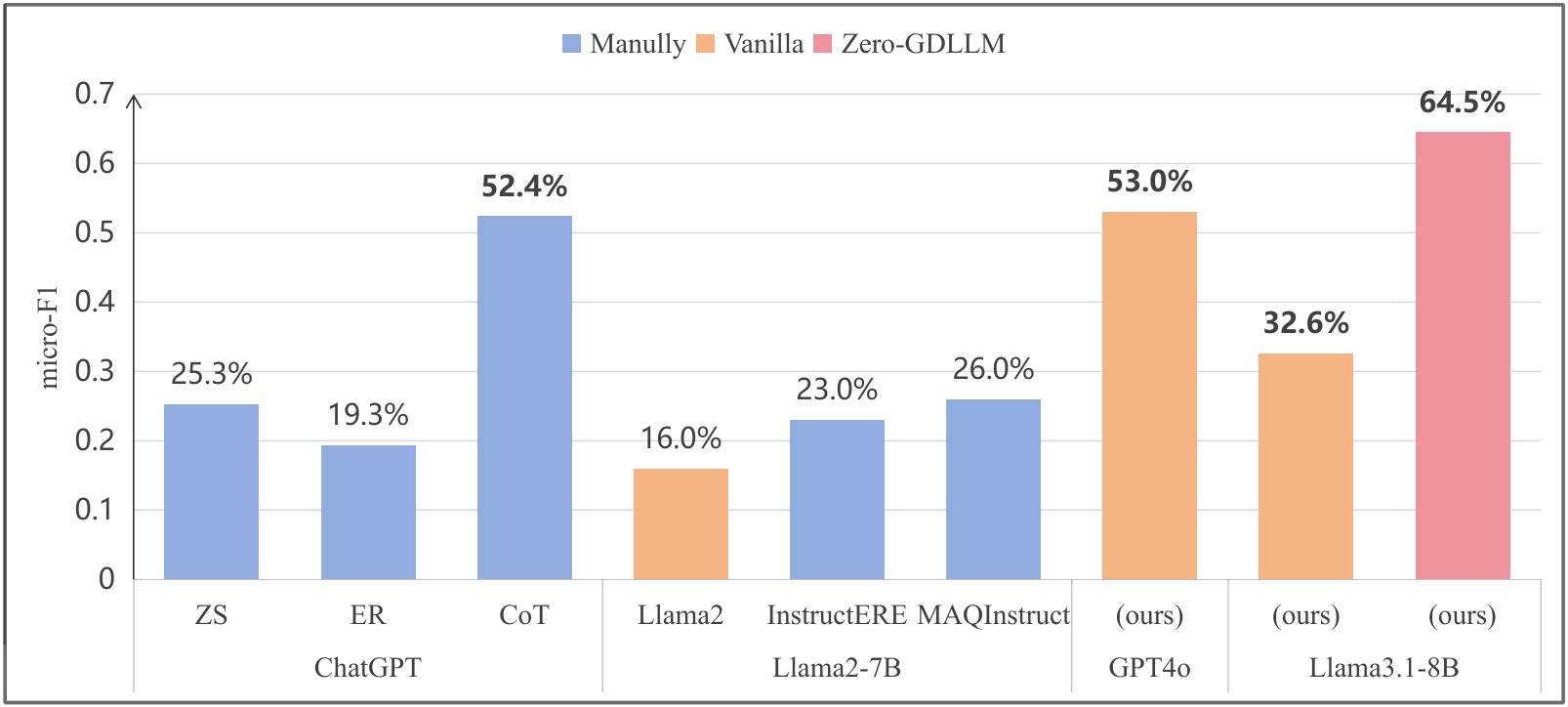}
  \caption{The micro-F1 score of the performance comparison on the MATRES dataset between our methods and other benchmarks based on zero-shot.}
  \label{fig:zero}
\end{figure}

As depicted in Figure~\ref{fig:zero}, we conduct various experiments to compare the zero-shot performance on the MATRES dataset with different benchmarks\citep{yuan:2023,xu:2025}. \textbf{1)For \textit{Manually} and \textit{Vanilla}}, \textit{Manually} means giving manually designed prompts or instructions to the ``\textit{Vanilla}'' LLMs which are not fine-tuned. Early work\citep{yuan:2023} designs three kinds of prompt techniques (ZS, ER, and CoT) to evaluate ChatGPT, which gives their best performance on the CoT prompts at 52.4\%. We report the result on vanilla GPT4o, which is higher than the CoT method. It suggests the importance of the scale of different LLMs and the limitations of manually designed prompts. \textbf{2)For \textit{Zero-GDLLM}}, it is to directly generate a probability distribution from Llama3.1 to GAT without LoRA tuning and the GAT operates with fixed parameters. We can see our \textit{Zero-GDLLM} method in Figure~\ref{fig:zero} outperforms all previous results above. That indicates the superior capacity of our distance-aware modeling approach in zero-shot learning scenarios.

For the input formats in zero-shot scenarios, we utilize hand-crafted prompts (e.g., ``I will give you a paragraph that uses \([EV1]\), \([/EV1]\), \([EV2]\) and \([/EV2]\) to, respectively mark two events, with the event relations divided into `\textit{BEFORE}', `\textit{AFTER}', `\textit{VAGUE}' and `\textit{EQUAL}'. You only need to provide the final judgment result of the event relation'') without task-specific training.



\begin{figure}[t]
\centering \includegraphics[width=0.95\columnwidth,height=0.9\textheight,keepaspectratio]{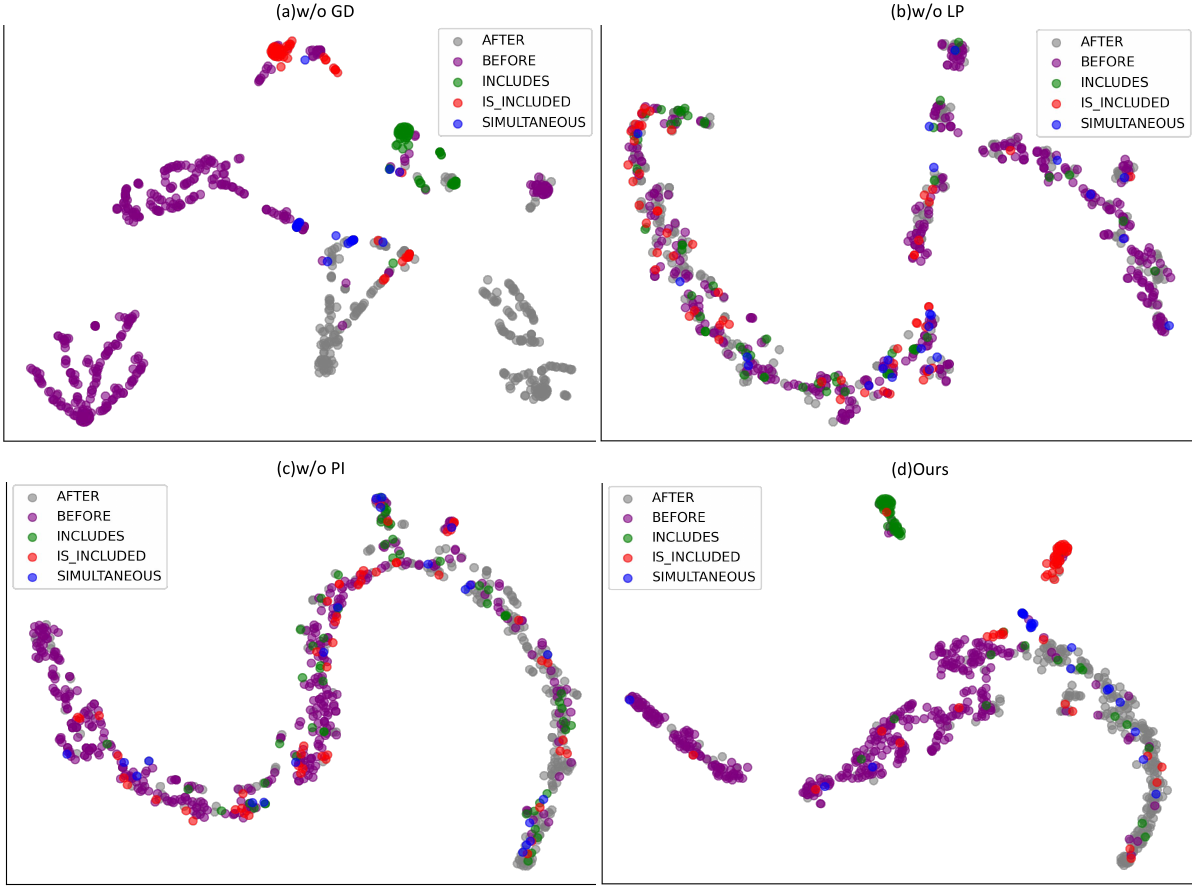}
  \caption{The visualized clustering comparison results of the ablation study based on Llama3.1-8B.}
  \label{fig:case}
\end{figure}

\subsection{Case Study for Minority Categories}

To evaluate the effectiveness of clustering minority categories, we visualize the final prediction result representations of positive samples in high-dimensional space. Specifically, we first obtain all representations on the testing set of the TB-dense dataset, which features highly imbalanced classes. Given the complex and non-linear nature of the data, we choose t-Distributed Stochastic Neighbor Embedding (t-SNE) as the dimension reduction technique to project the high-dimensional representations onto a two-dimensional space for visualization. We employ three baseline models following the ablation study.

As depicted in Figure~\ref{fig:case}(b) and Figure~\ref{fig:case}(c), the representation distribution of all positive examples has almost no obvious boundaries, which indicates the model performs poorly in clustering. Compared with Figure~\ref{fig:case}(a), it can be seen from Figure~\ref{fig:case}(d) that the t-SNE visualization of the proposed approach clearly separates and clusters minority relation classes, such as \textit{INCLUDES} and \textit{IS\_INCLUDED}, although there is still some minor overlap between classes, the distinct clustering patterns indicate that the model effectively captures the unique characteristics of these minority categories. This demonstrates that our approach effectively augments the capacity of capturing the global relation feature. Overall, the comparison results from the t-SNE visualization strongly demonstrate the superiority of the proposed model in handling minority temporal relation classes. 


\subsection{The Performance Comparison on GNN-based Benchmarks}
\label{sec:LLM}

\begin{figure}[t]
  \includegraphics[width=\linewidth]{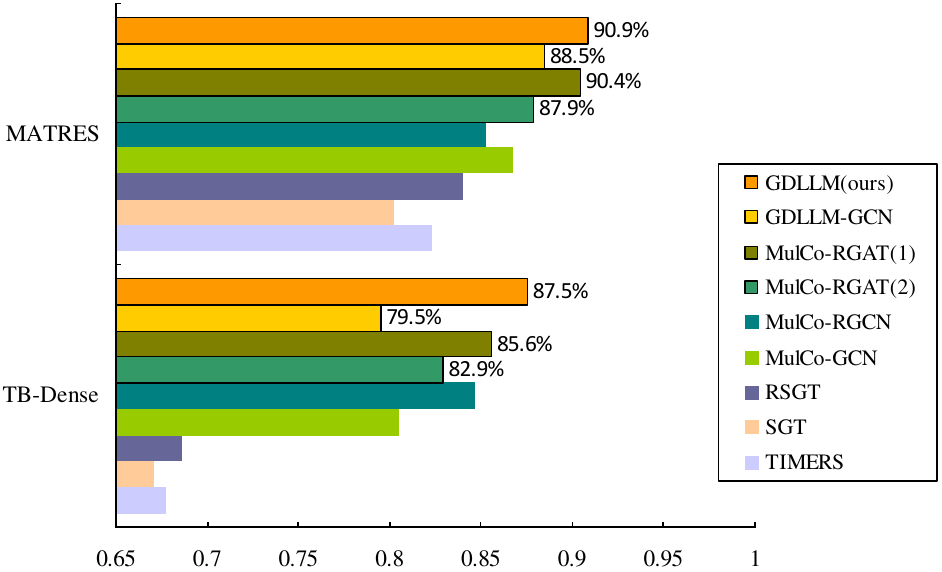}
  \caption{The micro-F1 score of the previous GNN-based method versus our approach on the two datasets. ``MulCo-RGAT(\(n\))'' represents the model adopts \(n\) GNN layers.}
  \label{fig:GNN}
\end{figure}

We analyze the performance of GNN-based methods with different benchmarks, which is depicted in Figure~\ref{fig:GNN}. The existing SOTA model MulCo\citep{yao:2024} contributes various GNN-based results. Our method, based on two layers of GAT, outperforms MulCo-RGAT(2), highlighting the effectiveness of our \textbf{GDLLM} proposed in the GNN-based approaches. We also test the performance of the GCN-based methods, which suggests that GCN lacks the capacity of multi-head attention, failing to effectively learn the probabilistic relation features for the short-distance proximity band.


\section{Related Work}

Earlier studies for ETRE predominantly rely on machine learning\citep{mani:2006,yoshikawa:2009}. Afterwards, some research integrates Pre-trained  Language Models to capture temporal semantics in the context \citep {cheng:2020,wen:2021,mathur:2021,man：2022}. It is also worth noting that more and more studies focus on the special structure of event temporal relations. One of the widely employed graph-based methods is GNNs. Different GNN-based methods have been proposed to better learn the relation cues \citep{mathur:2021,man：2022}. Differently, other researchers embed events in hyperbolic spaces for better hierarchical structure modeling\citep{tan:2021}. Prior SOTA model MulCo\citep{yao:2024} combines GNNs and the model of BERT variants via multi-scale knowledge distillation. 
There are also studies that tackle data scarcity or imbalance\citep{uzzaman:2013,wang:2020,han:2020,guan:2021,tan:2023,yuan:2024}, while some work designs certain temporal rules \citep{ballesteros:2020,zhuang:2023,ning:2024}. 

With the rapid development of LLMs, researchers pay great attention to the Question-Answer (QA) mechanism\citep{xu:2025,hu:2025}. Similar to the zero-shot studies, another work proposes a variety of valuable prompt explanations\citep{yuan:2023} or utilizes a unified framework\citep{huang:2023}. 

\section{Conclusion}
 In this paper, we propose \textbf{GDLLM}, a \textbf{G}lobal \textbf{D}istance-aware modeling approach based on \textbf{LLM}s. Specifically, we present a distance-aware graph structure
utilizing GAT to assist LLMs in capturing long-distance dependency features. Additionally, we design a temporal feature learning paradigm based on soft inference to augment the event relation extraction with a short-distance proximity band. Our framework also substantially enhances the performance of minority relation classes and improves the overall learning ability. Extensive experiments on two public datasets, TB-Dense and MATRES, demonstrate that our approach outperforms all LLM-based and GNN-based benchmarks, achieving SOTA performance without manually designed prompts or instructions for LLMs.

\section*{Limitations}

Although our method has already achieved the current state-of-the-art performance, the limitations may still exist. Due to the different category choices of LLMs, their inherent adaptability to task diversity or bias may pose challenges to our model training or performance. For example, on the minority class \textit{EQUAL}, the baseline utilizing the Qwen model exhibits suboptimal performance compared to the model CPTRE. Meanwhile, future work is needed to explore more effective and diverse modeling or training methods for Large Language Models.

\section*{Acknowledgments}
We would like to thank the anonymous reviewers for their helpful comments and suggestions.


\DeclareRobustCommand{\emph}[1]{\textit{#1}}

\DeclareRobustCommand{\uline}[1]{%
  \ifmmode #1\else 
  \expandafter\ifx\expandafter\textit\@car#1\@nil 
    #1 
  \else
    \uline@ #1\relax 
  \fi\fi
}
\bibliography{custom}

\appendix

\end{document}